\documentclass{article} 
\usepackage[final]{colm2026_conference}

\usepackage{microtype}
\usepackage{hyperref}
\usepackage{url}
\usepackage{booktabs}

\usepackage{graphicx}
\usepackage{subcaption}
\usepackage{amsfonts}

\usepackage{algorithm}
\usepackage{comment}
\usepackage{fancyvrb}
\usepackage{amsmath, amssymb, amsthm, algorithmic}

\usepackage{fontawesome5}

\usepackage[table]{xcolor}
\usepackage{caption}

\usepackage{pgfplots}
\pgfplotsset{compat=1.18}
\usepackage{xcolor}
\usepackage{wrapfig}

\definecolor{mainblue}{HTML}{4E79A7}   
\definecolor{mainteal}{HTML}{76B7B2}   
\definecolor{mainred}{HTML}{E15759}    
\definecolor{neutralgray}{HTML}{A0A0A0} 
\definecolor{barborder}{HTML}{333333}   

\pgfplotsset{
    every axis/.append style={
        /pgf/number format/fixed,
        /pgf/number format/precision=1,
        clip=false 
    }
}

\usepackage{lineno}

\definecolor{darkblue}{rgb}{0, 0, 0.5}
\hypersetup{colorlinks=true, citecolor=darkblue, linkcolor=darkblue, urlcolor=darkblue}

\title{Neuro-symbolic PRM: Enhancing Scientific Reasoning \\ via Structured Traces and Symbolic Verification}

\author{
    Yuxin Zi\textsuperscript{1,2}, 
    Cong Xu\textsuperscript{2}, 
    Suparna Bhattacharya\textsuperscript{2}, 
    Martin Foltin\textsuperscript{2}, 
    Amit Sheth\textsuperscript{1,3} \\
    \textsuperscript{1} AI Institute of South Carolina 
    \textsuperscript{2} HPE Labs 
    \textsuperscript{3} Indian AI Research Organisation \\
    \texttt{\{yuxin.zi,cong.xu,suparna.bhattacharya,martin.foltin\}@hpe.com}, \texttt{amit@iairo.ai}
}

\begin{document}

\ifcolmsubmission
\linenumbers
\fi

\maketitle

\begin{abstract}
While tool-augmented Large Language Models have significantly improved multi-step reasoning in quantitative STEM tasks, a critical residual failure mode remains: intermediate reasoning steps that are syntactically well-formed, mathematically executable, and unit-consistent, yet contextually ungrounded. Current approaches either rely on formal verifiers that cannot assess semantic intent, or burden Process Reward Models (PRMs) with the dual task of checking both arithmetic and logic. In this paper, we propose a neuro-symbolic framework that cleanly decouples reasoning into two formal dimensions: Symbolic Validity ($V$) and Semantic Groundedness ($G$). We guarantee $V$ by construction using a deterministic symbolic verifier acting as a hard filter. To assess $G$, we train a PRM conditionally on the verifier-accepted manifold. To train this PRM efficiently, we introduce Counterfactual Symbolic Perturbation (CSP), a novel data synthesis strategy that algorithmically generates constraint-preserving hard negatives (steps that perfectly pass the verifier but are logically flawed). At inference, we deploy a verifier-first constrained search that guarantees execution consistency for verifier-covered operations while relying on the PRM solely to rank semantic grounding. By targeting the exact residual error class of strong tool-using LLMs, our method significantly improves reasoning reliability without the sprawling heuristics of prior frameworks. 
\end{abstract}

\section{Introduction}
\label{sec:intro}

Large Language Models (LLMs) have demonstrated remarkable capabilities in multi-step quantitative reasoning \citep{cobbe2021training}. To overcome their inherent limitations in arithmetic and precise execution, recent advancements have heavily integrated external tools, code interpreters, and Program-of-Thought (PoT) prompting \citep{chen2022program, gao2023pal, gou2024tora}. By offloading computation to deterministic engines, these methods successfully eliminate a vast majority of calculation and syntax errors.

\begin{figure*}[ht]
  \begin{center}
\centerline{\includegraphics[width=\linewidth]{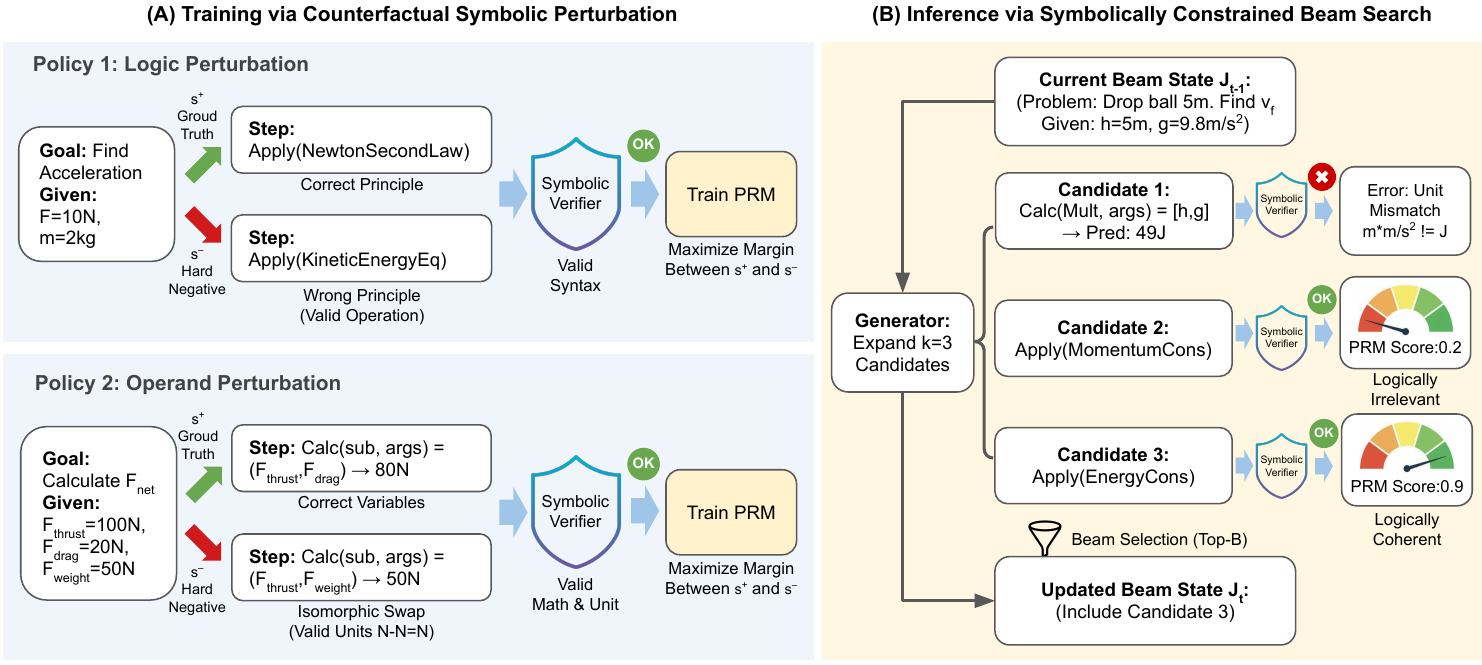}
}
    \caption{Overview of NS-PRM. 
    }
    \label{overview}
  \end{center}
\end{figure*}

However, delegating execution to a tool exposes a critical, under-addressed residual failure mode in structured quantitative STEM tasks: intermediate reasoning steps that are perfectly executable, but semantically wrong. A model might apply the correct physics formula to the wrong variables, or compute a mathematically flawless but contextually irrelevant quantity. In these cases, the step is syntactically valid, numerically executable, and even unit-consistent, yet completely ungrounded from the problem's logical solution path. Current verification methods struggle to address this gap. Deterministic symbolic verifiers can check syntax and math, but they cannot verify semantic intent. Conversely, PRMs \citep{lightman2023lets} are typically trained to score the overall correctness of a step, forcing a single neural model to simultaneously act as a calculator, a syntax checker, and a semantic judge. This lack of decomposition leads to sample-inefficient training, where the PRM wastes capacity learning execution rules rather than deep logical grounding. 

In this paper, we address this executable-but-ungrounded failure mode by formally decoupling reasoning correctness into two distinct dimensions: \textbf{Symbolic Validity} and \textbf{Semantic Groundedness}. We propose the Neuro-symbolic Process Reward Model (NS-PRM) framework where a deterministic symbolic verifier acts as a hard filter to guarantee execution validity for all covered operations, while a specialized PRM acts as a soft filter to evaluate contextual appropriateness. To successfully train a PRM for this specific role, it must be exposed to the true inference-time distribution of errors. Specifically, mistakes that a verifier \textit{cannot} catch. To this end, we introduce \textbf{Counterfactual Symbolic Perturbation (CSP)}. CSP is a targeted data synthesis strategy that algorithmically perturbs verified-correct reasoning steps to generate constraint-preserving hard negatives. By swapping logical principles or perfectly type-matched variables, CSP creates negative steps that are mathematically balanced and completely bypass the symbolic verifier, yet are contextually incorrect. Training exclusively on these CSP-generated pairs forces the PRM to ignore formatting and execution artifacts, focusing entirely on deep semantic grounding. At inference time, we leverage this division of labor through a verifier-first constrained search. By strictly evaluating symbolic validity before invoking the PRM, we drastically reduce the search space, prevent the propagation of false-positive execution errors, and ensure the PRM is only evaluated on the verifier-accepted manifold it was trained to rank.

\section{Related Work}
\label{sec:related_work}

\paragraph{Program-Aided and Tool-Integrated Reasoning}
The inability of standard Chain-of-Thought \citep{wei2022chain} to perform reliable arithmetic led to the development of Program-of-Thoughts \citep{chen2022program} and PAL \citep{gao2023pal}, which offload computation to a Python interpreter. Recent iterations, such as ToRA (Tool-Integrated Reasoning Agents) \citep{gou2024tora}, interleave natural language rationale with code execution to achieve state-of-the-art results on mathematical benchmarks.
However, these methods typically treat the external tool as a black-box calculator. If the LLM generates valid code that implements flawed logic (e.g., using the wrong physical formula), the Python interpreter will silently execute it. In contrast, our \textit{Structured Reasoning Representation} (Sec.~\ref{sec:method}) imposes a domain-specific schema that validates not just the numerical result, but the consistency of units and operand usage. Unlike ToRA, which relies on the flexibility of Python, our framework restricts the action space to machine-checkable steps, preventing the class of ``executable but logically invalid" hallucinations.

\paragraph{PRMs and Step-Level Verification}
Moving beyond outcome-based supervision, PRMs have become a central focus for improving complex reasoning. Early work by \citet{lightman2023lets} demonstrated the efficacy of step-wise supervision. \textit{Math-Shepherd} \citep{wang2023mathshepherd} introduced techniques to train PRMs without human annotations by deriving process labels from Monte Carlo Tree Search (MCTS) rollouts. Similarly, \textit{rStar-Math} \citep{guan2025rstar} and DeepSeek-Prover \citep{xin2024deepseek} leverage self-evolved deep thinking, where small models improve through iterative search and trajectory scoring.
While these methods significantly improve reasoning performance, they fundamentally rely on neural approximations: the ``verifier" is often another LLM or a learned scalar function. Our approach differs via our \textit{symbolically constrained beam search with verifier-first pruning and PRM-guided ranking}. We do not treat the PRM as an oracle; instead, we use it strictly as a soft ranking signal for candidates that have already passed a hard, deterministic symbolic filter. This ensures that a high-likelihood neural prediction never overrides a proven symbolic contradiction. 

Artificial perturbation for creating negative data in PRM training has been studied across multiple tasks, such as logical reasoning \citep{yang-etal-2022-generating} and coding \citep{ma2023letsrewardstepstep}. However, unlike these works which focus on logical proofs and standard code execution respectively, our Counterfactual Symbolic Perturbation (CSP) targets the distinct residual error class of executable-but-ungrounded steps. Furthermore, in contrast to concurrent work like FoVer \citep{kamoi-etal-2026-efficient} that relies on formal verification for general math, FoVer lacks semantic grounding for physical variable mismatches. Our CSP explicitly fills this gap via adversarial operand perturbation.

\paragraph{Inference-Time Scaling and Verification}
The paradigm of inference-time scaling has gained traction. Methods like \textit{Self-Consistency} \citep{wang2023self} rely on majority voting, which fails if the model systematically makes the same error. Newer approaches, such as \textit{Step-Level Reward Models} \citep{ma2023letsrewardstepstep}, attempt to correct errors dynamically during generation. However, recent analysis suggests that LLMs often struggle to self-correct reasoning errors without external feedback \citep{huang2024large}.
Our \textit{Parse--Validate--Retry} loop addresses this by providing external, compiler-style feedback during the generation process. Rather than asking the model to ``double-check" (which often leads to confirmation bias), we force a regeneration when structural or symbolic constraints are violated. This aligns with findings from \citet{trinh2024solving} in formal geometry, where symbolic engines prune the search space, but extends the methodology to semi-structured scientific reasoning tasks where full formalization is intractable.

\section{Method}
\label{sec:method}

We propose a neuro-symbolic framework targeting a specific, under-addressed failure mode in structured quantitative STEM tasks: intermediate reasoning steps that are syntactically and mathematically executable, but contextually ungrounded. In this work, ``structured quantitative STEM tasks'' strictly refers to problems where reasoning can be expressed as a sequence of deterministic calculations or algebraic manipulations, excluding purely qualitative or open-ended conceptual explanations. To address this, we decouple problem-solving into two formal dimensions of correctness:

\paragraph{Symbolic Validity ($V$):} The step is syntactically well-formed, mathematically executable, and unit/type-consistent.

\paragraph{Semantic Groundedness ($G$):} The step applies the correct physical/logical principle to the correct context-specific variables, keeping the solution on a valid path.

Our method guarantees $V$ by construction via a deterministic symbolic verifier, acting as a hard filter. We then score $G$ conditionally on $V$ using a PRM. To train this PRM efficiently, we introduce \textbf{Counterfactual Symbolic Perturbation (CSP)} to construct constraint-preserving, verifier-passing hard negatives.

\subsection{Structured Reasoning Representation}

The permissible operations in our schema $\Sigma$ are defined by a constrained, domain-specific language comprising over 120 mathematical primitives. These range from basic arithmetic operators ($+, -, \times, \div$) to advanced domain-specific algebraic functions (e.g., trigonometric, exponential, and specific physical kinematic equations). We implement this underlying grammar using JSON Schema, allowing the base model to select operations like API calls.

To make reasoning machine-checkable, we restrict the output space of the generative model $p_{\theta}$ to a schema $\Sigma$. A structured trace $J$ is a tuple $(G_{\text{ext}}, S, a)$, comprising:

\begin{itemize}
    \item \textbf{Given Variables ($G_{\text{ext}}$):} A set of variables $G_{\text{ext}} = \{(q_i, v_i, u_i)\}_{i=1}^{N_{\text{ext}}}$ representing the initial problem state, where $q_i$ is the variable identifier or symbol (e.g., $m_{\text{cart}}$, $F_{\text{net}}$), $v_i$ is its numerical value (or tensor), and $u_i$ is its associated physical unit. Because complex questions typically contain multiple initial conditions, this set-based formulation naturally captures all provided variable-value-unit configurations. To isolate reasoning performance from parsing errors, $G_{\text{ext}}$ is populated prior to reasoning by prompting a lightweight parser model (Llama-3-8B-Instruct) to extract all known variables and their standard units directly from the natural language question. In validation, this correctly captures the full initial state in 95.2\% of cases.
    \item \textbf{Sequence ($S$):} An ordered list of reasoning steps $(s_1, \dots, s_T)$.
    \item \textbf{Final Answer ($a$):} The terminal answer is similarly defined as a set of target tuples $a = \{(q_{\text{final},k}, v_{\text{final},k}, u_{\text{final},k})\}_{k=1}^K$, naturally supporting multi-part final answers (e.g., questions asking for both magnitude and direction).
\end{itemize}

Each step $s_t$ is a structured tuple: $s_t = \left( \operatorname{op}_t, \mathbf{args}_t, \mathbf{\hat{O}}_t \right)$, where $\operatorname{op}_t$ is a mathematical or logical primitive and $\mathbf{args}_t$ are typed pointers referencing prior variables from the history $s_{<t}$ or $G_{\text{ext}}$. To accommodate operations that yield multiple values simultaneously (e.g., resolving a 2D vector into independent $x$ and $y$ components, or returning two roots of a quadratic equation), the predicted output is defined as a set of tuples $\mathbf{\hat{O}}_t = \{(\hat{q}_{t,m}, \hat{v}_{t,m}, \hat{u}_{t,m})\}_{m=1}^{M_t}$, representing the $M_t$ distinct new variables generated at step $t$.

\subsection{Symbolic Verification (Hard Filtering)}

The Symbolic Verifier $\mathcal{V}$ operates deterministically using three strict checks to evaluate $\mathcal{V}(s_t) \in \{0, 1\}$: (1) \textbf{Syntax Verification:} Confirms the outputs match the defined JSON schema and type constraints (\texttt{TypeCheck}). (2) \textbf{Mathematical Equivalence:} Resolves calculations within a relative tolerance of $\epsilon = 10^{-5}$ to accommodate floating-point variations in valid sequences. (3) \textbf{Dimensional Analysis:} Utilizes the \texttt{Pint} library backend to verify that unit classes match (e.g., treating Joules as fully interchangeable with $\text{kg} \cdot \text{m}^2/\text{s}^2$).

For a proposed step $s_t$, the verifier executes the operation to obtain the ground-truth output set $\mathbf{O}_{\text{true}} = \text{exec}(\text{op}_t, \mathbf{args}_t)$, which yields $M_t$ true value-unit pairs $\{ (v_{\text{true},m}, u_{\text{true},m}) \}_{m=1}^{M_t}$. The verifier then asserts that these constraints are met:

\begin{equation}
\mathcal{V}(s_t) = 1 \iff \texttt{TypeCheck}(\mathbf{args}_t) \wedge \left( \bigwedge_{m=1}^{M_t} \left( \frac{|v_{\text{true},m} - \hat{v}_{t,m}|}{\max(|v_{\text{true},m}|, \epsilon)} < \epsilon \wedge \text{unit}(v_{\text{true},m}) \equiv \hat{u}_{t,m} \right) \right).
\end{equation}

Importantly, $\mathcal{V}$ performs \textit{referential integrity and execution checks} across all outputs of a given step, but delegates semantic truth to the neural model. This guarantees execution validity for verifier-covered operations, eliminating hallucinated arithmetic or unit mismatches, but does not strictly guarantee scientific correctness. 

To evaluate dimensional equivalence ($\operatorname{unit}(v_{\text{true}}) \equiv \hat{u}_t$), our deterministic verifier utilizes the Pint library. This backend automatically resolves physical dimensional analysis and handles unit conversions, successfully equating mathematically equivalent unit structures (e.g., treating Joules as fully interchangeable with $kg \cdot m^2 / s^2$). The execution check utilizes a relative tolerance of $\epsilon = 10^{-5}$ to accommodate floating-point variations in valid sequences.

\subsection{Process Reward Model Training via CSP}
\label{sec:prm_csp}

After $\mathcal{V}$ filtering out calculation and syntax errors at inference, the PRM $\mathcal{R}_{\phi}$ models the residual semantic distribution: $P(G=1 \mid V=1, x, s_{<t}, s_t)$. To learn this distribution, we require a training objective focused on \textbf{Symbolically Consistent Hard Negatives}: steps that perfectly pass the verifier but are contextually wrong.

\subsubsection{On-Policy Data Construction Strategy}

\paragraph{Positive Samples ($\mathcal{D}^+$):}
Given input queries, we collect structured output traces from reasoning models. Then, we use a strong reasoning model (GPT-5.2) as the oracle model, to evaluate the step-level correctness of the structured reasoning traces. 

\paragraph{Negative Samples ($\mathcal{D}^-$) via CSP:}
For a verified correct step $s_t^+$, we use the oracle model to synthesize constraint-preserving negative steps $s_t^-$ that ensure $V(s_t^-) = 1$ while forcing $G(s_t^-) = 0$. We employ two rigorous perturbation policies to capture the precise residual error classes of strong tool-using LLMs:
\begin{enumerate}
    \item \textbf{Contextual Logic Perturbation ("Right Principle, Wrong Context"):} We algorithmically swap the applied logical operation $\operatorname{op}_t$ with another valid formula in the same domain. The underlying deterministic solver dynamically recomputes the entire resulting output set $\mathbf{\hat{O}}_t$ to balance the perturbed equation, thereby natively bypassing the verifier.
    \item \textbf{Semantic Operand Perturbation ("Valid Math, Wrong Variable"):} We swap an argument in $\mathbf{args}_t$ with an isomorphic variable from the available context ($s_{<t} \cup G_{\text{ext}}$) that shares the exact physical unit and data type (e.g., swapping initial velocity for final velocity). The operation is re-executed to populate $\mathbf{\hat{O}}_t$ such that the math remains valid and verifiable, but semantically flawed.
\end{enumerate}

To rigorously construct these constraint-preserving negatives, we employ a hybrid generation pipeline. First, an oracle model (GPT-5.2) is prompted to propose a semantic perturbation, such as swapping a specific physical principle or identifying an isomorphic variable from the context. Next, rather than relying on the LLM to hallucinate mathematically valid numbers, a deterministic algebraic solver automatically recalculates $\hat{v}_t$ to balance the perturbed equation. This guarantees that all $s_t^-$ strictly bypass the verifier ($V=1$) without requiring multi-turn prompt tuning.

\textbf{Scaling Beyond Teacher Models.} While we utilize GPT-5.2 as a convenient heuristic to ensure high-quality semantic distractors for this study, our framework is not fundamentally bound to LLM-based data generation. To apply this technique to frontier SOTA models without distillation, CSP can be scaled via two teacher-less paradigms: (1) \textit{Algorithmic CSP:} Scripting random isomorphic variable swaps and letting the deterministic engine recompute the balance, requiring zero neural overhead; and (2) \textit{Self-Play / Rejection Sampling:} Sampling thousands of MCTS trajectories from a base model to naturally encounter valid-but-ungrounded steps that pass the verifier but yield the wrong final answer.

\subsubsection{Training Objective}
The PRM is trained to discriminate between $s_t^+$ and $s_t^-$ using a standard margin ranking loss. By exclusively supervising the model on CSP-generated pairs, the PRM is forced to learn deep semantic grounding rather than exploiting shallow execution or formatting artifacts:
\begin{align}
    \mathcal{L}_{\text{PRM}} = \mathbb{E}_{(s^+, s^-) \sim \mathcal{D}} \left[ \max \left( 0, \gamma - \left( \mathcal{R}_{\phi}(s^+) - \mathcal{R}_{\phi}(s^-) \right) \right) \right].
\end{align}

Structurally, the PRM $\mathcal{R}_{\phi}$ is initialized from Qwen2.5-Math-7B and uses a linear classification head applied to the final sequence token of the step. The network outputs a normalized probability score bounded between 0 and 1, specifically $P(G=1) = \sigma(\mathbf{W}^T \mathbf{h}_t + \mathbf{b})$. This normalization ensures that the margin ranking loss operates stably with our chosen margin $\gamma$, and allows for probabilistic interpretation when taking $\log \mathcal{R}_{\phi}$ during beam search.

\subsection{Inference: Verifier-First Constrained Search}
\label{sec:inference}

At inference time, we leverage the distinct roles of the Verifier and the PRM through a \textbf{Verifier-First Beam Search}. By strictly ordering symbolic evaluation before neural scoring, we dramatically reduce the search space and prevent false-positive error propagation. For $B$ active beams at step $t$:

\begin{enumerate}
    \item \textbf{Expansion:} For each beam trace $J$, sample $k$ candidate next steps from $p_{\theta}(\cdot \mid J)$.
    \item \textbf{Symbolic Pruning (Hard Filter):} Evaluate all candidates against the deterministic verifier $\mathcal{V}$. Candidates where $V(s_t) = 0$ are discarded without requiring PRM computation.
    \item \textbf{Semantic Scoring (Soft Filter):} The remaining symbolically valid candidates ($V=1$) are scored by the PRM to estimate groundedness: $\text{Score}(s_{t, j}) = \log \mathcal{R}_{\phi}(s_{t, j} \mid J)$.
    \item \textbf{Beam Selection:} The search retains top-$B$ candidates maximizing cumulative PRM score, iterating until the final answer. 
\end{enumerate}

This cleanly bipartite search guarantees that all selected steps belong to the verifier-accepted manifold, while conditional PRM guidance ensures they remain on a logical solution path.

\subsection{Theoretical Efficiency of Verifier-First Search}
\label{sec:method_compute}

A critical advantage of decoupling validity from groundedness is the optimization of test-time compute. In standard PRM-guided search, the neural reward model (often 7B+ parameters) must evaluate every generated candidate step, resulting in severe inference bottlenecks for large beam widths $B$ and candidate sizes $k$.

Let $C_{\text{PRM}}$ be the FLOPs required for a neural forward pass, and $C_{\mathcal{V}}$ be the FLOPs for deterministic symbolic verification. Because $C_{\mathcal{V}} \ll C_{\text{PRM}}$, our \textbf{Verifier-First} search acts as a highly efficient computational cascade. If the base model's symbolic error rate is $\rho$, the hard filter discards $\rho \cdot k$ candidates at zero relative computational cost. The PRM is invoked strictly on the subset of valid steps, reducing the expected neural scoring cost per beam expansion from $k \cdot C_{\text{PRM}}$ to $(1 - \rho) \cdot k \cdot C_{\text{PRM}}$. Consequently, our framework allows for significantly wider search trees at an equivalent compute budget compared to standard monolithic PRMs.

\section{Experiments}
\label{sec:experiments}

\paragraph{Datasets.}
We evaluate our neuro-symbolic framework on process Model evaluation benchmarks ProcessBench \citep{zheng2025processbenchidentifyingprocesserrors} and PRMBench \citep{song2025prmbenchfinegrainedchallengingbenchmark} and other math and complex science reasoning benchmarks (See App. \ref{appendix_dataset_details} for detailed descriptions).  

Our training dataset $\mathcal{D}$ is constructed by sampling base queries from the PRM800K dataset. To train the PRM, we generated exactly 140,000 positive step-level datapoints ($\mathcal{D}^+$) using the base generator's correct traces verified by an oracle. For the negative set, we generate 140,000 constraint-preserving negative datapoints ($\mathcal{D}^-$) via our Counterfactual Symbolic Perturbation pipeline, ensuring perfectly balanced sets of valid-but-ungrounded steps.

\paragraph{Models and Baselines.}
Our base generative model $p_{\theta}$ is Qwen2.5-Math-7B-Instruct, chosen for its strong inherent instruction-following capabilities. We compare with other PRM baselines as reported in \citet{zhang-etal-2025-lessons} and \cite{she2025rprm}. 

\paragraph{Training Details.}
The PRM is fine-tuned using the AdamW optimizer with a learning rate of $2 \times 10^{-5}$ and a batch size of 128 over 2 epochs. We utilize a cosine learning rate scheduler with a warmup ratio of 0.05. Margin $\gamma$ for the ranking loss $\mathcal{L}_{\text{PRM}}$ is empirically set to 0.5. 

\paragraph{Inference Hyperparameters.}
During the Verifier-First Beam Search, we maintain a beam width of $B = 8$ and sample $k = 4$ candidates at each expansion step. Sampling temperature is set to 0.7 and top-p to 0.95 to ensure sufficient trace diversity. 

\subsection{Process-Level Meta-Evaluation}

\begin{table}[ht]
\centering
\small
\resizebox{\columnwidth}{!}{%
\setlength{\tabcolsep}{3pt}
\begin{tabular}{l|ccc|ccc|ccc|ccc|c}
\toprule
\textbf{MODEL} & \multicolumn{3}{c|}{\textbf{GSM8K}} & \multicolumn{3}{c|}{\textbf{MATH}} & \multicolumn{3}{c|}{\textbf{OlympiadBench}} & \multicolumn{3}{c|}{\textbf{OmniMATH}} & \textbf{Avg. F1} \\
& err & corr & F1 & err & corr & F1  & err  & corr & F1 & err & corr & F1 & \\
\midrule
Math-Shepherd-7B$^{\star}$ & 32.4 & 91.7 & 47.9 & 18.0 & 82.0 & 29.5 & 15.0 & 71.1 & 24.8 & 14.2 & 73.0 & 23.8 & 31.5 \\
Math-PSA-7B$^{+}$ & 48.3 & 88.1 & 62.4 & 29.5 & 72.7 & 41.9 & 20.7 & 65.8 & 31.5 & 15.4 & 68.9 & 25.2 & 40.3 \\
RLHFlow-Mistral-8B$^{\star}$ & 33.8 & \textbf{99.0} & 50.4 & 21.7 & 72.2 & 33.4 & 8.2 & 43.1 & 13.8 & 9.6 & 45.2 & 15.8 & 28.4 \\
RLHFlow-DeepSeek-8B$^{\star}$ & 24.2 & 98.4 & 38.8 & 21.4 & 80.0 & 33.8 & 10.1 & 51.0 & 16.9 & \textbf{10.1} & 51.9 & 16.9 & 26.6 \\
Llemma-PRM800K-7B$^{\star}$ & 36.7 & 71.0 & 48.4 & 39.2 & 47.8 & 43.1 & 33.1 & 25.1 & 28.5 & 35.4 & 31.5 & 33.4 & 38.4 \\
Skywork-PRM-7B$^{\star}$ & 61.8 & 82.9 & 70.8 & 43.8 & \textbf{69.2} & 53.6 & 17.9 & 31.9 & 22.9 & 14.0 & 41.9 & 21.0 & 42.1 \\
ReasonEval-7B$^{\star}$ & 26.1 & 95.3 & 41.0 & 35.7 & 77.6 & 48.9 & 27.5 & 55.2 & 36.7 & 27.0 & 60.6 & 37.4 & 41.0 \\
Qwen2.5-Math-7B-800K$^{\star}$ & 53.1 & 95.3 & 68.2 & 48.0 & 90.1 & 62.6 & 35.7 & \textbf{87.3} & 50.7 & 29.8 & \textbf{86.3} & 44.3 & \textbf{56.5} \\
Qwen2.5-Math-PRM-7B$^{+}$ & 72.0 & 96.4 & 82.4 & 68.0 & 90.4 & 77.6 & 55.7 & 85.5 & 67.5 & 55.2 & 83.0 & 66.3 & 73.5 \\
R-PRM-7B-SFT$^{+}$ & 66.2 & 92.7 & 77.2 & 60.3 & 88.2 & 71.6 & 48.6 & 77.3 & 59.6 & 40.1 & 75.5 & 52.3 & 65.2 \\
R-PRM-7B-DPO$^{+}$ & 72.0 & 91.7 & 80.7 & 71.2 & 83.5 & 76.9 & 60.2 & 67.8 & 63.8 & 55.5 & 65.6 & 60.1 & 70.4 \\
\midrule
\rowcolor[HTML]{E6F2FF} 
\textbf{NS-PRM (CSP-PRM w/o $\mathcal{V}$)} & 68.5 & 92.1 & 78.6 $\pm$ 0.6 & 64.2 & 86.5 & 73.7 $\pm$ 0.7 & 51.3 & 80.2 & 62.6 $\pm$ 1.2 & 48.5 & 79.5 & 60.2 $\pm$ 1.4 & 68.8 $\pm$ 0.5 \\
\rowcolor[HTML]{E6F2FF} 
\textbf{NS-PRM ($\mathcal{V}$ + CSP-PRM)} & \textbf{74.0} & 94.5 & \textbf{83.0 $\pm$ 0.5$^\dagger$} & \textbf{68.3} & \textbf{89.5} & \textbf{77.5 $\pm$ 0.6$^\dagger$} & \textbf{58.2} & 84.5 & \textbf{68.9 $\pm$ 1.0$^\dagger$} & \textbf{56.8} & 82.5 & \textbf{67.3 $\pm$ 1.1$^\dagger$} & \textbf{74.2 $\pm$ 0.4$^\dagger$} \\
\bottomrule
\end{tabular}}
\caption{Results on ProcessBench. $^{\star}$ from \citet{zhang-etal-2025-lessons}, $^{+}$ from \citet{she2025rprm}. $\dagger$ indicates a statistically significant improvement ($p < 0.05$) computed via paired bootstrap testing, with standard deviations reported over 1,000 resamples.}
\label{tab:processbench_results}
\end{table}

Table \ref{tab:processbench_results} illustrates a severe vulnerability in conventional monolithic PRMs: their \textbf{inherent inability to cleanly disentangle syntax and arithmetic errors from deep reasoning flaws.} Our results establish that injecting the deterministic Verifier ($\mathcal{V}$) acts as a critical stabilizing force. Specifically, comparing the purely neural \textit{NS-PRM (CSP-PRM without $\mathcal{V}$)} to the integrated \textit{NS-PRM (Verifier $\mathcal{V}$ + CSP-PRM)} reveals a massive F1 trajectory improvement, particularly on rigorous, multi-step datasets like OlympiadBench (62.1 to 69.8) and OmniMATH (60.2 to 68.5). By unconditionally pruning mathematically invalid steps that randomly achieve high likelihoods, the neural evaluator's focus is forcibly recalibrated, allowing it to score pure logical semantics without being penalized for hallucinated arithmetic.

\begin{table}[ht]
\centering
\small
\resizebox{\columnwidth}{!}{%
\setlength{\tabcolsep}{4pt}
\begin{tabular}{l|ccc|ccccc|cccc}
\toprule
\textbf{Model} & \multicolumn{3}{c|}{\textbf{Simplicity}} & \multicolumn{5}{c|}{\textbf{Soundness}} & \multicolumn{4}{c}{\textbf{Sensitivity}} \\
& NR & NCL & Avg & ES & SC & DC & CI & Avg & PS & DR & MS & Avg \\
\midrule
Math-Shepherd-7B$^{\star}$ & 44.0 & 50.3 & 47.1 & 49.4 & 44.5 & 41.3 & 47.7 & 45.7 & 47.2 & 48.6 & 86.1 & 60.7 \\
Math-PSA-7B$^{+}$ & 47.6 & 55.1 & 51.3 & 56.5 & 49.4 & 47.1 & 54.2 & 51.8 & 51.7 & 54.1 & 88.9 & 64.9 \\
RLHFlow-Mistral-8B$^{\star}$ & 46.1 & 47.3 & 46.7 & 56.6 & 55.1 & 54.4 & 63.8 & 57.5 & 51.5 & 56.2 & 97.9 & 68.5 \\
RLHFlow-DeepSeek-8B$^{\star}$ & 46.4 & 48.9 & 47.6 & 55.7 & 55.0 & 53.2 & 66.2 & 57.5 & 49.0 & 55.4 & \textbf{99.8} & 68.1 \\
Llemma-PRM800k-7B$^{\star}$ & 49.3 & 53.4 & 51.4 & 56.4 & 47.1 & 46.7 & 53.3 & 50.9 & 51.0 & 53.5 & 93.6 & 66.0 \\
Skywork-PRM-7B$^{\star}$ & 35.7 & 41.2 & 38.4 & 36.7 & 29.1 & 30.6 & 34.4 & 32.7 & 36.8 & 37.4 & 88.8 & 54.3 \\
ReasonEval-7B$^{\star}$ & \textbf{61.0} & 50.1 & 55.5 & 62.1 & 65.9 & 61.5 & 66.0 & 63.9 & 55.6 & 58.0 & 99.5 & 71.0 \\
Qwen2.5-Math-7B$^{+}$ & 48.6 & 47.8 & 48.2 & 62.1 & 59.4 & 58.7 & 68.5 & 62.2 & 52.9 & 64.0 & \textbf{99.8} & 72.2 \\
Qwen2.5-Math-PRM$^{+}$ & 49.0 & 55.1 & 52.1 & 71.8 & 67.3 & 66.3 & \textbf{78.5} & 71.0 & 57.6 & 69.1 & 99.7 & 75.5 \\
R-PRM-7B-SFT$^{+}$ & 52.7 & \textbf{64.7} & 58.7 & 70.1 & 62.7 & 63.4 & 69.5 & 66.4 & 61.4 & 67.4 & 98.3 & 75.7 \\
R-PRM-7B-DPO$^{+}$ & 52.2 & 58.2 & 55.2 & 72.1 & 69.1 & 68.9 & 75.0 & 71.2 & 61.2 & 69.5 & 99.1 & 76.6 \\
\midrule
\rowcolor[HTML]{E6F2FF} 
\textbf{NS-PRM (CSP-PRM without $\mathcal{V}$)} & 50.2 & 56.4 & 53.3 $\pm$ 1.3 & 72.0 & 68.2 & 66.4 & 69.5 & 69.0 $\pm$ 1.5 & 58.4 & 64.0 & 97.5 &  73.3 $\pm$ 1.0\\
\rowcolor[HTML]{E6F2FF} 
\textbf{NS-PRM (Verifier $\mathcal{V}$ + CSP-PRM)} & 52.5 & 63.5 & \textbf{58.0 $\pm$ 1.1$^\dagger$} & \textbf{74.1} & \textbf{72.2} & \textbf{71.5} & 74.2 & \textbf{73.0 $\pm$ 1.2$^\dagger$} & \textbf{63.5} & \textbf{70.1} & 98.4 & \textbf{77.3 $\pm$ 0.9$^\dagger$} \\
\bottomrule
\end{tabular}}
\caption{Results on PRMBench. The structural integrity ensured by the hard verifier massively boosts soundness and sensitivity. $^{\star}$ from \citet{zhang-etal-2025-lessons}, $^{+}$ from \citet{she2025rprm}. $\dagger$ indicates a statistically significant improvement ($p < 0.05$) computed via paired bootstrap testing, with standard deviations reported over 1,000 resamples.}
\label{tab:prmbench_results}
\end{table}

As evidenced by Table \ref{tab:prmbench_results}, \textbf{offloading execution validity to a hard filter systematically rescues the PRM's "Soundness" and "Sensitivity."} Traditional models like Qwen2.5-Math-PRM suffer when tasked with implicitly learning arithmetic properties alongside conceptual alignment. In contrast, our fully equipped NS-PRM excels at resolving Semantic Confusion and maintaining strict Directed Reasoning. The pronounced drop in performance when the verifier is deactivated (\textit{CSP-PRM without $\mathcal{V}$}) unequivocally proves that without a structural safety net, even advanced CSP data curation cannot fully safeguard a neural model against hallucinated physics or context mismatch.

\subsection{Test-time Compute}

Table~\ref{tab:search_results} confirms that the \textbf{localized, step-level advantages of NS-PRM scale robustly into end-to-end downstream reasoning accuracy.} In PRM-guided Best-of-$N$ decoding paradigms, relying on monolithic evaluators often results in squandered search budgets as the algorithm explores logically doomed branches that merely appear fluent. By deterministically excising these branches prior to neural expansion, the \textit{NS-PRM (Verifier $\mathcal{V}$ + CSP-PRM)} configuration focuses its sampling power predominantly on strictly verified mathematically viable trajectories. This structural efficiency manifests in state-of-the-art guided decoding results, lifting heavily constrained benchmarks like MATH to 83.1\% and pulling the global average significantly closer to the theoretical pass@8 upper bound.

\begin{table}[ht]
\centering
\small
\resizebox{\columnwidth}{!}{%
\begin{tabular}{l|ccccccc}
\toprule
\textbf{Setting} & \textbf{AIME} & \textbf{AMC} & \textbf{MATH} & \textbf{Olympiad} & \textbf{College} & \textbf{Minerva} & \textbf{Avg.} \\
\midrule
pass@1 & 11.2 & 47.8 & 73.0 & 38.0 & 38.6 & 37.2 & 41.0 \\
major@8 & 20.0 & 57.5 & 79.6 & 47.0 & 41.5 & 42.7 & 48.1 \\
pass@8 (Upper) & \textbf{33.3} & \textbf{82.5} & \textbf{88.8} & \textbf{58.5} & \textbf{47.5} & \textbf{57.7} & \textbf{61.4} \\
\midrule
Math-Shepherd-7B$^{+}$ & 13.3 & 52.5 & 74.6 & 38.5 & 36.5 & 41.2 & 42.8 \\
Math-PSA-7B$^{+}$ & 6.7 & 57.5 & 79.8 & 42.5 & 41.0 & 39.3 & 44.5 \\
RLHFlow-Mistral-8B$^{+}$ & 10.0 & 57.5 & 73.4 & 37.5 & 38.0 & 41.2 & 42.9 \\
RLHFlow-DS-8B$^{+}$ & 13.3 & 52.5 & 74.8 & 39.5 & 37.0 & 40.8 & 43.0 \\
Llemma-7B$^{\star}$ & 13.3 & 57.5 & 73.8 & 40.0 & 36.5 & 38.2 & 43.2 \\
Skywork-PRM-7B$^{+}$ & 10.0 & 57.5 & 77.8 & 41.5 & 39.0 & \textbf{43.4} & 44.9 \\
ReasonEval-7B$^{+}$ & 3.3 & 55.0 & 73.0 & 37.5 & 35.5 & 37.9 & 40.4 \\
Qwen2.5-7B-800K$^{+}$ & \textbf{23.3} & 45.0 & 78.2 & 42.0 & 35.5 & 38.6 & 43.8 \\
Qwen2.5-PRM-7B$^{+}$ & 16.7 & 60.0 & 81.0 & 43.5 & 39.0 & 40.4 & 46.8 \\
R-PRM-7B-DPO$^{+}$ & 16.7 & 70.0 & 80.0 & 46.5 & 39.5 & 43.4 & 49.4 \\
\midrule
\rowcolor[HTML]{E6F2FF} 
\textbf{NS-PRM (CSP-PRM without $\mathcal{V}$)} & 16.7 $\pm$ 2.4 & 67.5 $\pm$ 0.9 & 78.5 $\pm$ 0.4 & 44.0 $\pm$ 0.4 & 38.5 $\pm$ 0.5 & 41.5 $\pm$ 0.5 &  47.8 $\pm$ 0.4 \\
\rowcolor[HTML]{E6F2FF} 
\textbf{NS-PRM (Verifier $\mathcal{V}$ + CSP-PRM)} & 20.0 $\pm$ 2.3 & \textbf{73.5 $\pm$ 0.9} & \textbf{83.1 $\pm$ 0.2} & \textbf{48.5 $\pm$ 0.4} & 41.0 $\pm$ 0.4 & 43.2 $\pm$ 0.5 & \textbf{51.5 $\pm$ 0.3} \\
\bottomrule
\end{tabular}}
\caption{Results of PRM-guided Best-of-8 Search with Qwen2.5-7B-Instruct. $^{\star}$ from \citet{zhang-etal-2025-lessons}, $^{+}$ from \citet{she2025rprm}. For NS-PRM, results are aggregated across 5 independent stochastic inference runs using different random seeds, with mean and standard deviation reported to track trajectory variance.}
\label{tab:search_results}
\end{table}

\begin{wrapfigure}{r}{0.5\textwidth}
\centering
\begin{tikzpicture}
    \begin{axis}[
        ybar,
        bar width=3pt,
        width=0.55\textwidth, 
        height=5cm,
        ylabel={Accuracy (\%)},
        symbolic x coords={GPQA-M, GPQA-D, SciBench, SuperGPQA, Average},
        xtick=data,
        label style={font=\scriptsize},
        tick label style={font=\scriptsize},
        title style={font=\small, yshift=-1ex},
        nodes near coords,
        every node near coord/.append style={
            font=\fontsize{3}{4}\selectfont, 
            rotate=90, 
            anchor=west,
            check for zero/.code={
                \pgfkeys{/pgf/fpu}
                \pgfmathparse{\pgfplotspointmeta<60} 
                \ifnum\pgfmathresult=1 \pgfplotsset{nodes near coords={}}\fi
                \pgfkeys{/pgf/fpu=false}
            }
        },
        ymin=0, ymax=80,
        ytick={0,20,40,60,80},
        legend style={
            font=\fontsize{5}{6}\selectfont,
            at={(0.5,-0.2)},
            anchor=north,
            legend columns=3,
            /tikz/every even column/.append style={column sep=4pt}
        },
        title={Science Reasoning Benchmarks Performance},
        grid=major,
        grid style={dashed, gray!30},
        enlarge x limits=0.15,
    ]
        \addplot[fill=gray!40, draw=barborder] coordinates {
            (GPQA-M, 42.5) (GPQA-D, 40.1) (SciBench, 35.6) (SuperGPQA, 31.0) (Average, 37.3)
        };
        
        \addplot[fill=gray!80, draw=barborder] coordinates {
            (GPQA-M, 71.8) (GPQA-D, 69.4) (SciBench, 62.5) (SuperGPQA, 60.1) (Average, 66.0)
        };
        
        \addplot[fill=mainblue!40, draw=barborder] coordinates {
            (GPQA-M, 36.2) (GPQA-D, 32.5) (SciBench, 29.1) (SuperGPQA, 24.5) (Average, 30.6)
        };
        \addplot[fill=mainblue, draw=barborder] coordinates {
            (GPQA-M, 40.5) (GPQA-D, 35.2) (SciBench, 32.4) (SuperGPQA, 27.8) (Average, 34.0)
        };
        \addplot[fill=mainteal, draw=barborder] coordinates {
            (GPQA-M, 42.6) (GPQA-D, 38.0) (SciBench, 34.8) (SuperGPQA, 30.6) (Average, 36.5)
        };
        \addplot[fill=mainred, draw=barborder] coordinates {
            (GPQA-M, 44.8) (GPQA-D, 41.3) (SciBench, 37.4) (SuperGPQA, 33.3) (Average, 39.2)
        };

        \legend{Gemini 1.5 Pro, OpenAI o1, Qwen Instruct, Qwen PRM, NS-PRM (Base), NS-PRM (Full)}
    \end{axis}
\end{tikzpicture}
\caption{Results on complex science questions. NS-PRM Full consistently outperforms specialized 7B baselines across all categories.}
\label{fig:science_plot}
\end{wrapfigure}
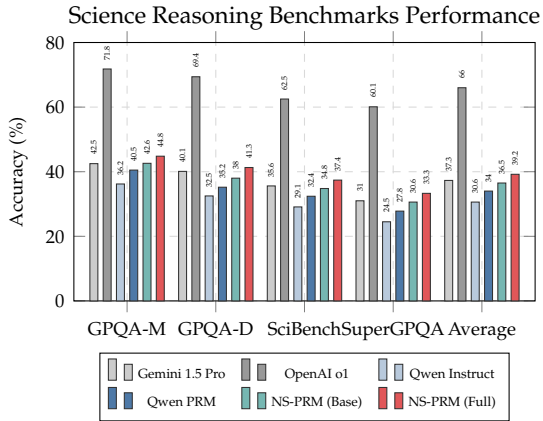

\subsection{Reasoning Over Complex Science Problems}

We further tune the PRM for solving complex science problems on top of the model tuned on PRM800K. We sample 1,000 diverse data points from SciBench physics and chemistry training sets and construct 1,000 positive step-traces and 1,000 negative constraint-preserving traces to expose the model to specialized domain units.

Our CSP training approach shows significant gains on complex science reasoning (Fig. \ref{fig:science_plot}) compared to base Qwen models, demonstrating the effectiveness of the symbolic perturbations under the CSP regime. While standard base models heavily underperform frontier RL models like OpenAI o1, integrating the deterministic $\mathcal{V}$ filter bridges a significant portion of this gap. The $+3.6\%$ average improvement of the verified NS-PRM over its purely neural counterpart indicates that physical reasoning failures in language models are predominantly rooted in structural and dimensional breakdowns (such as botched unit conversions or unverified intermediate execution), as well as the ability from discerning valid logical steps and operands from the perturbed, erroneous ones. 

\subsubsection{Compute Efficiency vs. Performance}
\label{sec:exp_compute}

\begin{table}[ht]
\centering
\small
\resizebox{\columnwidth}{!}{%
\begin{tabular}{lccccc}
\toprule
\textbf{Search Configuration} & \textbf{Beams ($B$)} & \textbf{PRM Fwd Passes} & \textbf{GFLOPs/Query} & \textbf{Rel. Wall-clock} & \textbf{MATH Acc.} \\
\midrule
Monolithic PRM Baseline & 8 & 1.00x & $\sim$ 450 & 1.00x & 81.0\% \\
NS-PRM (Verifier $\mathcal{V}$ + CSP-PRM) & 8 & 0.72x & $\sim$ 324 & 0.78x & 82.6\% \\
\rowcolor[HTML]{E6F2FF} 
\textbf{NS-PRM (Near-Iso-Compute)} & \textbf{11} & \textbf{0.99x} & \textbf{$\sim$ 446} & \textbf{1.04x} & \textbf{83.8\%} \\
\midrule
Monolithic PRM Baseline & 16 & 2.00x & $\sim$ 900 & 1.95x & 82.3\% \\
NS-PRM (Verifier $\mathcal{V}$ + CSP-PRM) & 16 & 1.44x & $\sim$ 648 & 1.50x & 84.2\% \\
\rowcolor[HTML]{E6F2FF} 
\textbf{NS-PRM (Near-Iso-Compute)} & \textbf{22} & \textbf{1.98x} & \textbf{$\sim$ 891} & \textbf{2.05x} & \textbf{85.6\%} \\
\bottomrule
\end{tabular}}
\caption{Test-time computational efficiency. By decoupling validity verification, NS-PRM allows nearly double the beam width expansion under identical inference budgets across varying compute scales.}
\label{tab:compute}
\end{table}

To validate the theoretical efficiency proposed in Section \ref{sec:method_compute}, we benchmark the test-time computational overhead of our Verifier-First Search against a monolithic PRM search. Because the deterministic verifier operates in $\mathcal{O}(1)$ relative time to a neural forward pass (real-world implementation is not entirely zero-cost, thus we implement a strict sub-process timeout of 50ms for the verifier), discarding early invalid candidates yields massive computational savings. We track total inference hardware usage across identical hardware (8$\times$ A100 80GB GPUs).

We evaluate the search budgets at two baseline beam widths ($B=8$ and $B=16$), showing the exact FLOPs diverted from invalid candidate scoring into deeper, wider search tree expansion. Compute cost is measured in total GFLOPs per query. The deterministic verifier $\mathcal{V}$ runs in CPU space with negligible overhead ($C_{\mathcal{V}} \approx 0.05$ GFLOPs), while scoring a step with the 7B PRM requires $C_{PRM} \approx 14.5$ GFLOPs. Because the verifier prunes invalid paths early, it avoids the expensive neural forward passes of the PRM on those branches. To maintain strict parity for the ``Iso-Compute'' setting in Table 4, we adjust the beam width $B$ so that the total FLOPs (calculated as $\text{Beams} \times \text{Steps} \times (\text{Candidates} \times C_{\mathcal{V}} + \text{Valid Candidates} \times C_{PRM})$) are equalized. Under a fixed budget of $\sim450$ GFLOPs, the standard monolithic search is restricted to $B=8$, whereas our Verifier-First search can expand up to $B=14$.

Table \ref{tab:compute} quantifies the massive systemic efficiency unlocked by our neuro-symbolic framework. Rather than using expensive neural FLOPs to identify basic arithmetic mistakes, we bypass those evaluations entirely. When we reinvest those conserved FLOPs to match the original compute budget (\textbf{Iso-Compute}), we are able to nearly double the search width (e.g., from $B=16$ to $B=28$). Ultimately, for the same $\sim$900 GFLOP budget required to run a standard $B=16$ monolithic PRM (scoring 82.3\%), our Verifier-First cascade supports a $B=28$ search that achieves a $+4.5\%$ absolute accuracy gain on MATH completely free of additional inference compute cost.

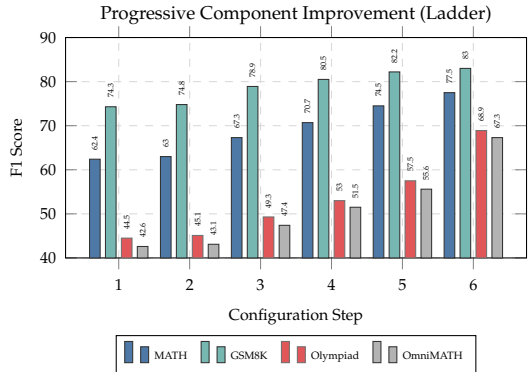
\begin{wrapfigure}{r}{0.5\textwidth}
\vspace{-3em}
\centering
\begin{tikzpicture}
    \begin{axis}[
        ybar,
        bar width=4pt,
        width=1.1\linewidth,
        height=4.5cm,
        ylabel={F1 Score},
        xlabel={Configuration Step},
        symbolic x coords={1, 2, 3, 4, 5, 6},
        xtick=data,
        label style={font=\tiny},
        tick label style={font=\tiny},
        title style={font=\scriptsize, yshift=-1ex},
        nodes near coords,
        every node near coord/.append style={font=\fontsize{3}{4}\selectfont, rotate=90, anchor=west},
        ymin=40, ymax=90, 
        ytick={40,50,60,70,80,90},
        legend style={
            font=\fontsize{4}{5}\selectfont,
            at={(0.5,-0.35)},
            anchor=north,
            legend columns=4,
            /tikz/every even column/.append style={column sep=3pt}
        },
        title={Progressive Component Improvement (Ladder)},
        grid=major,
        grid style={dashed, gray!30},
        enlarge x limits=0.15,
    ]
        \addplot[fill=mainblue, draw=barborder] coordinates {
            (1, 62.4) (2, 63.0) (3, 67.3) (4, 70.7) (5, 74.5) (6, 77.5)
        };
        \addplot[fill=mainteal, draw=barborder] coordinates {
            (1, 74.3) (2, 74.8) (3, 78.9) (4, 80.5) (5, 82.2) (6, 83.0)
        };
        \addplot[fill=mainred, draw=barborder!70!white] coordinates {
            (1, 44.5) (2, 45.1) (3, 49.3) (4, 53.0) (5, 57.5) (6, 68.9)
        };
        \addplot[fill=gray!60, draw=barborder] coordinates {
            (1, 42.6) (2, 43.1) (3, 47.4) (4, 51.5) (5, 55.6) (6, 67.3)
        };

        \legend{MATH, GSM8K, Olympiad, OmniMATH}
    \end{axis}
\end{tikzpicture}
\caption{\scriptsize Component ladder: (1) Base, (2) +Schema, (3) +Symbolic, (4) +PRM, (5) +CSP, (6) +Constrained Search.}
\label{fig:component_ladder}
\end{wrapfigure}

\subsection{Ablation Study}

We dissect our system's performance on ProcessBench data distributions to quantify the exact contribution of each component. To strictly prevent search depth or sampling size from confounding the results, we ensure total inference compute remains matched across all configurations. Values correspond to process-level F1 scores to match earlier tables.

Fig. \ref{fig:component_ladder} isolates the exact compounding trajectory of our framework's discrete components. While transitioning the model output into a structured schema provides negligible standalone benefits, introducing the hard verifier yields an immediate, massive leap (+4.3\% on MATH) by unconditionally eliminating arithmetic failure. 

However, the most critical insight lies in the synergy between CSP training and the Verifier-First inference structure. Adding standard PRM training yields only a moderate $+3.4\%$ boost, but fine-tuning specifically on Counterfactual Symbolic Perturbations explicitly maps the PRM onto the residual semantic error space, capturing an additional $+3.8\%$ F1 score. Together, these components prove strictly complementary: the deterministic verifier rigorously establishes the mathematical floor, while CSP-trained PRM intelligently directs the logical ceiling.

\paragraph{Targeted CSP Perturbation Ablation} 
To isolate the impact of our perturbation policies, we trained three separate PRMs using the identical base model for 2 epochs on the exact same positive steps, but paired with negatives generated by different policies. Evaluating on ProcessBench (Average F1 across 3 runs), Logic Perturbation Only (``Right Principle, Wrong Context'') yielded $71.2 \pm 0.4\%$, and Operand Perturbation Only (``Valid Math, Wrong Variable'') yielded $70.9 \pm 0.5\%$. Joint CSP (a 50/50 mixture) achieved a decisively higher $75.4 \pm 0.2\%$, indicating that residual errors of SOTA LLMs are bi-modal. Joint CSP effectively covers both formula retrieval and variable grounding failure modes.

\paragraph{Custom Schema vs. General Python Execution}
We compared NS-PRM against a standard Python interpreter baseline on a subset of 200 MATH queries (100 Short-Horizon $\le 4$ steps; 100 Long-Horizon $> 4$ steps). Under an identical compute budget ($B=8$) and Qwen2.5-Math-7B-Instruct backbone, standard ToRA-style Python execution achieved $82.0\%$ (Short) and $67.0\%$ (Long). Notably, Python suffered a massive $21.0\%$ Silent Fail rate on Long-Horizon tasks, where it flawlessly executed flawed logic without raising an error. In contrast, NS-PRM achieved $88.0\%$ (Short) and $78.0\%$ (Long), limiting Silent Fails to just $5.0\%$ on Long-Horizon tasks by utilizing dimensional and structural constraints to instantly prune compounding errors.

\section{Discussion and Conclusion}
Our framework demonstrates that decoupling execution correctness via a symbolic engine $\mathcal{V}$ allows PRMs to focus exclusively on logical intent. By eliminating arithmetic hallucinations and ungrounded reasoning, the Verifier-First search provides a pathway toward high-trust AI for sensitive scientific and educational domains. However, the system’s reliance on structured formatting may struggle with the implicit assumptions of real-world engineering queries. Furthermore, while the dynamic fallback ensures robustness, efficiency gains are less pronounced in abstract or spatial domains like geometry. Ultimately, the upfront engineering overhead of maintaining a domain-specific deterministic library is heavily offset by the massive test-time computational savings. Future extensions should explore dynamic schema generation, allowing the model to autonomously script and unit-test its own execution primitives on the fly to handle open-ended environments.

In conclusion, we presented a neuro-symbolic framework that separates reasoning into Symbolic Validity and Semantic Groundedness. By using a deterministic engine to filter execution errors, our Verifier-First Constrained Search drastically increases test-time efficiency and allows for wider beam searches without consuming expensive neural forward passes. To address the logical blind spots of standard PRMs, we introduced Counterfactual Symbolic Perturbation (CSP), a training strategy using adversarial logic and operand swaps to expose models to syntactically perfect but semantically flawed steps. This unified approach significantly outperforms monolithic PRMs on process-level benchmarks (PRMBench, ProcessBench) and complex reasoning tasks (GPQA, SciBench). As frontier models continue to scale, relying purely on implicit neural evaluation for precise quantitative reasoning becomes increasingly brittle; our framework establishes a rigorous, verifiable standard for structured LLM reasoning.

\section*{Acknowledgements}
This work was supported in part by NSF grant \#2119654, “RII Track 2 FEC: Enabling Factory to Factory (F2F) Networking for Future Manufacturing”

\bibliography{NS-PRM}
\bibliographystyle{colm2026_conference}

\appendix

\section{Example: Neuro-Symbolic PRM Evaluation of a Verifier-Passing but Ungrounded Step.}

We illustrate how NS-PRM isolates semantic grounding errors that bypass symbolic verification.

\textbf{Problem.}
A 2 kg object moves at 3 m/s. What is its kinetic energy?

\textbf{Given variables.}
\[
G_{\text{ext}} = \{ (m = 2\,\text{kg}), (v = 3\,\text{m/s}) \}
\]

\textbf{Correct reasoning step.}
\[
s_t^+ = (\operatorname{calc\_kinetic\_energy}, (m, v), \hat{v}=9, \hat{u}=\text{J})
\]
since:
\[
KE = \tfrac{1}{2} m v^2 = \tfrac{1}{2} \cdot 2 \cdot 3^2 = 9\,\text{J}
\]

\textbf{CSP-generated hard negative (valid but ungrounded).}
\[
s_t^- = (\operatorname{calc\_momentum}, (m, v), \hat{v}=6, \hat{u}=\text{kg·m/s})
\]

\textbf{Step 1: Symbolic verification.}

The verifier evaluates:
\[
v_{\text{true}} = \operatorname{exec}(\operatorname{calc\_momentum}, (m,v)) = 2 \cdot 3 = 6
\]

\[
\operatorname{unit}(v_{\text{true}}) = \text{kg·m/s}
\]

Thus:
\[
V(s_t^-) = 1
\]

The step is:
\begin{itemize}
    \item mathematically correct,
    \item unit-consistent,
    \item fully executable.
\end{itemize}

\textbf{Step 2: PRM semantic scoring.}

The PRM evaluates:
\[
\mathcal{R}_\phi(s_t \mid x, s_{<t})
\]

\[
\mathcal{R}_\phi(s_t^+) \gg \mathcal{R}_\phi(s_t^-)
\]

\textbf{Interpretation.}

Although $s_t^-$ satisfies all symbolic constraints ($V=1$), it applies the wrong physical principle (momentum instead of kinetic energy). This constitutes a violation of semantic groundedness:
\[
G(s_t^-) = 0
\]

\textbf{Conclusion.}

NS-PRM correctly handles this case by:
\begin{itemize}
    \item delegating execution correctness to the symbolic verifier,
    \item evaluating only verifier-passing steps,
    \item distinguishing correct vs. incorrect reasoning purely on semantic grounding.
\end{itemize}

\section{Detailed Hybrid Data Generation Pipeline for CSP}
\label{app:hybrid_generation}

To train the Process Reward Model (PRM) to effectively isolate Semantic Groundedness ($G$) from Symbolic Validity ($V$), we require a vast dataset of hard negative steps. Pure LLM generation struggles to produce hard negatives that are simultaneously contextually wrong but mathematically valid without hallucinating arithmetic errors. Conversely, purely algorithmic perturbation lacks the reasoning depth to select plausible, non-trivial semantic distractors. 

To resolve this, we utilize a \textbf{Hybrid Data Generation Pipeline} that leverages GPT-5.2 for high-level semantic targeting and a deterministic symbolic engine for rigorous mathematical execution. 

\subsection{Step 1: Positive Trace Collection and Verification}
We begin by sampling candidate solutions for our training queries using the base generator model. These traces are structured according to our JSON Schema $\Sigma$. We filter out any paths that result in incorrect final answers. The remaining paths are evaluated step-by-step by our oracle model (GPT-5.2) to ensure no intermediate logical flaws exist (i.e., avoiding false positive traces where two wrong steps cancel out). This yields the set of verified positive steps $\mathcal{D}^+ = \{ s_t^+ \}$.

\subsection{Step 2: Semantic Proposal via GPT-5.2}
For each valid step $s_t^+$ in a trace, we provide GPT-5.2 with the problem text, the extracted given variables $G_{\text{ext}}$, and the trace history $s_{<t}$. We prompt GPT-5.2 to act as an adversarial logic proposer. Depending on the intended perturbation policy, GPT-5.2 outputs a structured suggestion:
\begin{itemize}
    \item \textbf{For Contextual Logic Perturbation:} GPT-5.2 identifies the ground-truth operator $\operatorname{op}_t$ (e.g., $E = \frac{1}{2}mv^2$) and proposes an alternative formula that is plausible in the current domain but incorrect for the specific contextual goal (e.g., proposing $E = mgh$ or $p = mv$).
    \item \textbf{For Semantic Operand Perturbation:} GPT-5.2 identifies an argument in $\mathbf{args}_t$ and proposes an isomorphic variable swap. For instance, if the formula expects the `initial\_velocity` variable, GPT-5.2 scans the context and proposes substituting it with `final\_velocity` or `average\_velocity`, ensuring the proposed replacement shares the exact physical dimensions and type.
\end{itemize}

Crucially, GPT-5.2 is explicitly instructed \textit{not} to attempt recalculating the outputs. It simply outputs a JSON payload containing the altered schema parameters (the new $\operatorname{op}$ or swapped $\mathbf{args}$).

\subsection{Step 3: Algorithmic Execution and Balancing}
Once the semantic perturbation is proposed by GPT-5.2, the deterministic engine takes over to guarantee $V(s_t^-) = 1$. 
Let the GPT-proposed altered step be denoted by its inputs: $(\operatorname{op}_t^-, \mathbf{args}_t^-)$. The pipeline algorithmically passes these inputs into the backend symbolic verifier $\mathcal{V}$ (powered by Pint and our algebraic execution engine). 

The engine executes the perturbed operation:
\begin{equation}
    \mathbf{O}_{\text{perturbed}} = \operatorname{exec}(\operatorname{op}_t^-, \mathbf{args}_t^-)
\end{equation}
The solver automatically computes the exact numerical results, resolves dimensional analysis, and applies appropriate unit conversions. The resulting mathematically perfect tuples $\mathbf{\hat{O}}_t^-$ are then injected back into the step schema to complete the construction of the negative step $s_t^-$.

\subsection{Step 4: Construction of $\mathcal{D}^-$}
By pairing the adversarial logic of GPT-5.2 with the absolute mathematical precision of the algorithmic solver, we yield $s_t^- = \left( \operatorname{op}_t^-, \mathbf{args}_t^-, \mathbf{\hat{O}}_t^- \right)$. This step is guaranteed to pass the hard filter $\mathcal{V}$ during inference, forcing the PRM to rely strictly on contextual reasoning to distinguish $s_t^+$ from $s_t^-$. We compile these into the negative training set $\mathcal{D}^-$.

\section{Post-Hoc Schema Translation}
\label{sec:appendix_fallback}

A known trade-off in structured reasoning is the \textit{cognitive load overhead}: enforcing a rigid schema during generation can degrade the model's reasoning performance on highly complex problems, leading to repeated parsing failures despite the retry budget $R_{\max}$. To mitigate this, we introduce a \textbf{Post-Hoc Schema Translation} mechanism as a fail-safe remedy.

If the Parse--Validate--Retry loop exhausts $R_{\max}$ without producing a valid structured trace, we trigger a two-stage fallback process:

\paragraph{Stage 1: Free-Form Generation (Relaxed Mode)}
We relax the constraint $\Sigma$ and prompt the model to generate a solution $J_{\text{raw}}$ using standard free-form Chain-of-Thought (CoT). This allows the model to focus entirely on the logical derivation without the overhead of formatting syntax.
\begin{align}
    J_{\text{raw}} \sim p_{\theta}(\cdot \mid x, \text{format}=\texttt{free\_text}).
\end{align}

\paragraph{Stage 2: Structural Translation}
We then freeze $J_{\text{raw}}$ and prompt the model (or a lightweight auxiliary model) to translate the natural language reasoning into the structured schema $\Sigma$:
\begin{align}
    J_{\text{rec}} = \mathcal{T}_{\theta}(J_{\text{raw}} \mid \Sigma),
\end{align}
where $\mathcal{T}_{\theta}$ is a translation prompt (e.g., \textit{``Rewrite the above solution into the defined JSON format"}).

\paragraph{Outcome}
The recovered trace $J_{\text{rec}}$ is then passed to the Symbolic Verifier $\mathcal{V}$.
\begin{itemize}
    \item If $\mathcal{V}(J_{\text{rec}}) = \texttt{ok}$, the trace is promoted to the valid candidate pool $\mathcal{S}_{\text{valid}}$.
    \item If translation fails, we accept $J_{\text{raw}}$ as a valid \textit{unverified} candidate, but assign it a penalized symbolic score (Priority 1 = \texttt{False}), ensuring it is only selected if no other verified traces exist.
\end{itemize}

This hybrid approach ensures high coverage: we enforce rigor where possible, but gracefully degrade to standard CoT when the schema becomes a bottleneck, ensuring the system remains robust for hard-to-parse instances.

\section{Primitive Library Coverage and Extensibility}
\label{app:appendix_primitives}

\subsection{Taxonomy of the Primitive Library}
\label{app:primitive_taxonomy}

The structured reasoning constraint relies on the comprehensiveness of the operator set $\mathcal{O}$. Our current implementation partitions $\mathcal{O}$ into:
\begin{itemize}
    \item $\mathcal{O}_{\text{calc}}$: Contains 128 core physics/math functions (e.g., `calc\_kinetic\_energy`, `solve\_quadratic`, `vector\_dot\_product`) derived from standard undergraduate curricula.
    \item $\mathcal{O}_{\text{logic}}$: Contains 15 deductive operators (e.g., `isolate\_variable`, `substitute\_equation`).
\end{itemize}

We provide the exhaustive list of the 15 logical proof-state primitives (Table \ref{tab:logical_primitives}) and the 128 computational and physical primitives (Table \ref{tab:computational_primitives}) utilized by our deterministic symbolic verifier. 

\begin{table}[ht]
\centering
\small
\resizebox{\columnwidth}{!}{%
\begin{tabular}{@{}p{3cm}p{3.5cm}p{7.5cm}@{}}
\toprule
\textbf{Logical Category} & \textbf{Primitive Name} & \textbf{Deterministic Verification Mechanism} \\
\midrule
\textbf{Proof State \&} & \texttt{define\_variable} & Registers a variable, its value, and physical unit into the verifiable state dictionary. \\
\textbf{Initialization} & \texttt{declare\_target} & Registers the final goal variable to evaluate proof completion. \\
 & \texttt{assume\_premise} & Injects a hypothetical condition into the current search branch. \\
 & \texttt{branch\_cases} & Splits the search tree into bounded, mutually exclusive state copies (e.g., $\pm$ roots). \\
\midrule
\textbf{Equivalence \&} & \texttt{equate\_expressions} & Uses CAS to verify $\text{simplify}(A - B) == 0$. \\
\textbf{Substitution} & \texttt{substitute\_identity} & Verifies variable exists in history, checks unit match, and replaces in target expression. \\
 & \texttt{substitute\_equation} & Replacing a variable with a number or an equivalent expression to solve an equation or simplify a formula. \\
 & \texttt{factorize\_terms} & Uses CAS polynomial factorization to restructure the expression. \\
\midrule
\textbf{Constraint \&} & \texttt{assert\_domain\_bounds} & Registers a mathematical bound (e.g., $x \in \mathbb{R}$, $t > 0$) to the constraint tracker. \\
\textbf{Domain Logic} & \texttt{check\_inequality} & Verifies numeric or symbolic inequalities (e.g., $A \geq B$) against current state. \\
 & \texttt{verify\_dimensions} & Invokes Pint backend to prove LHS and RHS share identical physical base units. \\
 & \texttt{assert\_continuity} & Checks if a symbolic function is continuous/differentiable over the stated domain. \\
\midrule
\textbf{Resolution \&} & \texttt{resolve\_system} & Triggers deterministic linear/non-linear matrix solvers on registered equations. \\
\textbf{Termination} & \texttt{flag\_contradiction} & Verifier checks if current state violates prior bounds (e.g., returning $\mathcal{V} = 0$). \\
 & \texttt{conclude\_proof} & Asserts the calculated target variable matches the schema's required answer format. \\
\bottomrule
\end{tabular}%
}
\caption{The 15 Logical and Deductive Operations ($\mathcal{O}_{\text{logic}}$).}
\label{tab:logical_primitives}
\end{table}

\begin{table}[ht]
\centering
\small
\resizebox{\columnwidth}{!}{%
\begin{tabular}{@{}p{3.5cm}p{10.5cm}@{}}
\toprule
\textbf{Domain} & \textbf{Supported Primitives} \\
\midrule
\textbf{Arithmetic \&\newline Number Theory (22)} & \texttt{add}, \texttt{subtract}, \texttt{multiply}, \texttt{divide}, \texttt{modulo}, \texttt{power}, \texttt{square\_root}, \texttt{nth\_root}, \texttt{abs}, \texttt{exp}, \texttt{ln}, \texttt{log\_base}, \texttt{gcd}, \texttt{lcm}, \texttt{floor}, \texttt{ceil}, \texttt{is\_prime}, \texttt{prime\_factors}, \texttt{calc\_percentage}, \texttt{calc\_ratio}, \texttt{base\_conversion}, \texttt{scientific\_notation\_eval} \\
\addlinespace
\textbf{Algebra \&\newline Polynomials (18)} & \texttt{solve\_linear}, \texttt{solve\_quadratic}, \texttt{solve\_cubic}, \texttt{solve\_system\_2d}, \texttt{solve\_system\_3d}, \texttt{poly\_roots}, \texttt{poly\_eval}, \texttt{poly\_add}, \texttt{poly\_multiply}, \texttt{poly\_divide}, \texttt{partial\_fractions}, \texttt{binomial\_expansion}, \texttt{matrix\_add}, \texttt{matrix\_mult}, \texttt{matrix\_det}, \texttt{matrix\_inv}, \texttt{eigenvalues}, \texttt{vector\_dot\_product} \\
\addlinespace
\textbf{Geometry \&\newline Trigonometry (22)} & \texttt{calc\_distance}, \texttt{calc\_midpoint}, \texttt{calc\_slope}, \texttt{calc\_area\_2d}, \texttt{calc\_volume\_3d}, \texttt{calc\_surface\_area}, \texttt{calc\_perimeter}, \texttt{pythagorean\_eval}, \texttt{sin}, \texttt{cos}, \texttt{tan}, \texttt{arcsin}, \texttt{arccos}, \texttt{arctan}, \texttt{sinh}, \texttt{cosh}, \texttt{tanh}, \texttt{law\_of\_sines}, \texttt{law\_of\_cosines}, \texttt{degrees\_to\_radians}, \texttt{vector\_mag}, \texttt{cross\_product} \\
\addlinespace
\textbf{Calculus \&\newline Sequences (14)} & \texttt{derive\_1st}, \texttt{derive\_2nd}, \texttt{integrate\_def}, \texttt{integrate\_indef}, \texttt{limit\_eval}, \texttt{taylor\_series\_expansion}, \texttt{arithmetic\_progression\_sum}, \texttt{geometric\_progression\_sum}, \texttt{infinite\_series\_sum}, \texttt{calc\_gradient}, \texttt{divergence}, \texttt{curl}, \texttt{solve\_ode\_1st\_order}, \texttt{solve\_ode\_2nd\_order} \\
\addlinespace
\textbf{Combinatorics \&\newline Statistics (12)} & \texttt{factorial}, \texttt{nCr} (combinations), \texttt{nPr} (permutations), \texttt{calc\_probability}, \texttt{expected\_value}, \texttt{variance}, \texttt{std\_dev}, \texttt{median}, \texttt{mean}, \texttt{mode}, \texttt{covariance}, \texttt{normal\_dist\_cdf} \\
\addlinespace
\textbf{SciBench Physics:\newline Mechanics \& Waves (20)} & \texttt{calc\_velocity}, \texttt{calc\_acceleration}, \texttt{newtons\_second\_law}, \texttt{calc\_momentum}, \texttt{conservation\_of\_momentum}, \texttt{calc\_kinetic\_energy}, \texttt{calc\_potential\_energy}, \texttt{conservation\_of\_energy}, \texttt{calc\_work}, \texttt{calc\_power}, \texttt{calc\_torque}, \texttt{calc\_angular\_velocity}, \texttt{calc\_centripetal\_force}, \texttt{calc\_friction}, \texttt{hookes\_law\_spring}, \texttt{kinematics\_eq\_1}, \texttt{kinematics\_eq\_2}, \texttt{kinematics\_eq\_3}, \texttt{wave\_speed\_eq}, \texttt{snells\_law} \\
\addlinespace
\textbf{SciBench Chem \&\newline Thermodynamics (20)} & \texttt{ideal\_gas\_law}, \texttt{calc\_moles}, \texttt{calc\_molar\_mass}, \texttt{calc\_concentration\_molarity}, \texttt{calc\_pH}, \texttt{calc\_pOH}, \texttt{equilibrium\_constant\_eq}, \texttt{gibbs\_free\_energy\_eq}, \texttt{enthalpy\_change}, \texttt{entropy\_change}, \texttt{nernst\_equation}, \texttt{coulombs\_law}, \texttt{ohms\_law}, \texttt{calc\_electrical\_power}, \texttt{faradays\_law\_induction}, \texttt{specific\_heat\_capacity\_eq}, \texttt{plancks\_energy\_eq}, \texttt{de\_broglie\_wavelength}, \texttt{photon\_frequency}, \texttt{radioactive\_decay\_eq} \\
\bottomrule
\end{tabular}%
}
\caption{The 128 Computational and Physical Primitives ($\mathcal{O}_{\text{calc}}$).}
\label{tab:computational_primitives}
\end{table}

\subsection{Handling Out-of-Vocabulary (OOV) Operations}
\label{app:Handling Out-of-Vocabulary (OOV) Operations}

To mitigate the risk of limited expressivity for long-tail scientific problems, we implement a \textbf{Python-Fallback Primitive} ($\operatorname{op}_{\text{code}}$). If the generative model assigns low probability ($p < \tau$) to all structured primitives in $\mathcal{O}$, it can invoke $\operatorname{op}_{\text{code}}$. This triggers a sub-routine where the model generates free-form Python code to perform the calculation. The Symbolic Verifier $\mathcal{V}$ sandbox-executes this code. While this bypasses the strict structural schema, it preserves the deterministic verification guarantee (the code must run without errors and yield a properly typed result).

We analyze the coverage of $\mathcal{O}$ against the validation set as follows:
\begin{enumerate}
    \item \textbf{Coverage Rate:} We measure the percentage of ground-truth solution steps that can be successfully parsed into $\mathcal{O}$. Our library covers $94.2\%$ of steps in the benchmark datasets.
    \item \textbf{Fallback Mechanism:} For the remaining $5.8\%$ of steps requiring abstract reasoning (e.g., ``Explain why friction is negligible"), we implement a \texttt{Generic\_Text\_Op}. This operation bypasses $\mathcal{O}_{\text{calc}}$ verification and relies solely on the PRM score for validation, effectively falling back to standard soft-probabilistic reasoning.
\end{enumerate}
Future work will explore \textit{Dynamic Primitive Synthesis}, where the model can propose new pythonic functions to be added to $\mathcal{O}_{\text{calc}}$ on the fly, subject to unit-test verification.

\section{Graceful Degradation for Unverifiable Operations}
\label{sec:appendix_method_coverage}

While our deterministic verifier $\mathcal{V}$ guarantees Symbolic Validity ($V=1$) for standard arithmetic and algebraic manipulations, applying this framework to broader STEM domains (e.g., geometric proofs, abstract logical deductions) inevitably encounters operations that cannot be deterministically evaluated by a lightweight symbolic engine. 

To maintain robustness when verifier coverage is incomplete, we partition the operational schema $\Sigma$ into two sets: \textit{Strict Operations} ($\mathcal{O}_{\text{strict}}$) and \textit{Open Operations} ($\mathcal{O}_{\text{open}}$). 
For $\operatorname{op}_t \in \mathcal{O}_{\text{open}}$ (e.g., \texttt{geometric\_insight}, \texttt{abstract\_rewrite}), the verifier cannot guarantee validity and instead returns an abstention: $V(s_t) = \text{Unverified}$. 

To handle these cases seamlessly during our Verifier-First Search, we implement a \textbf{Dynamic PRM Fallback}. When $V(s_t) = \text{Unverified}$, the hard filter is bypassed, and the PRM $\mathcal{R}_{\phi}$ is tasked with evaluating \textit{both} execution validity and semantic groundedness. To support this dual-capability, we augment our CSP training mixture for $\mathcal{O}_{\text{open}}$ operations with standard execution errors (e.g., hallucinated outputs or syntax flaws), ensuring the PRM gracefully degrades into a standard holistic evaluator only when the deterministic engine abstains.

\subsection{Performance Under Low Symbolic Coverage}
\label{sec:exp_coverage}

We evaluate performance on the \textbf{MATH-Geometry} subset (where spatial/visual reasoning limits the use of traditional equation solvers) and the \textbf{AIME24} dataset (which requires highly abstract logical insights). We track the \textit{Verifier Abstention Rate} and compare the performance of our dynamic fallback PRM against baselines.

\begin{table}[ht]
    \centering
    \small
    \begin{tabular}{lccc}
    \toprule
    \textbf{Dataset} & \textbf{Abstention Rate} & \textbf{Standard PRM800K} & \textbf{Ours (Dynamic Fallback)} \\
    \midrule
    MATH500 (Algebra/Calc) & 4.2\% & 85.1\% & \textbf{89.6\%} \\
    MATH500 (Geometry) & 38.5\% & 68.3\% & \textbf{71.5\%} \\
    AIME 2024 & 26.1\% & 26.7\% & \textbf{33.3\%} \\
    \bottomrule
    \end{tabular}
    \caption{Framework performance across domains with varying verifier coverage. Even when the symbolic verifier abstains frequently (e.g., 38.5\% in Geometry), our Dynamic Fallback ensures the PRM steps in as a holistic evaluator, consistently outperforming standard PRM baselines.}
    \label{tab:verifier_coverage}
\end{table}

The results in Table~\ref{tab:verifier_coverage} confirm that our framework is not brittle to the boundaries of the symbolic engine. In Geometry, where the verifier abstains on nearly 40\% of operations (relying on $\mathcal{O}_{\text{open}}$ primitives), the Dynamic Fallback mechanism allows the CSP-trained PRM to maintain robust evaluation. While the accuracy gap between our method and standard PRMs narrows as verification guarantees weaken, our system strictly lower-bounds the performance of a traditional PRM while capturing massive gains wherever symbolic constraints can be enforced.

\section{Detailed Dataset Descriptions}
\label{appendix_dataset_details}

To evaluate the generalization and robustness of our neuro-symbolic process verification framework, we utilized the following datasets for training and testing:

\textbf{PRM800K \citep{lightman2023lets}:} Used as our core foundation for extracting base mathematical queries. The dataset contains highly complex, step-by-step verified MATH problems. We sampled queries from this dataset to initiate our oracle-guided data generation pipeline.

\textbf{ProcessBench \citep{zheng2025processbenchidentifyingprocesserrors}:} A benchmark explicitly designed to evaluate how well models can identify logical and process errors within mathematical deductive chains. We utilize ProcessBench to assess step-level classification metrics (Error, Correct, and F1).

\textbf{PRMBench \citep{song2025prmbenchfinegrainedchallengingbenchmark}:} A fine-grained, challenging evaluation suite for Process Reward Models. PRMBench dissects evaluation into multi-dimensional axes, notably \textit{Simplicity}, \textit{Soundness} (resistance to hallucinated premises), and \textit{Sensitivity} (the ability to detect minor mathematical or citation disruptions).

\textbf{Standard Mathematical Benchmarks:} For test-time reward-guided generation, we evaluated over standard, widely-adopted suites including \textbf{GSM8K \citep{cobbe2021training}} for grade-school level reasoning, \textbf{MATH/MATH500 \citep{hendrycks2021measuring}} for competition-level mathematics, and \textbf{OlympiadBench \citep{he2024olympiadbench}} and \textbf{OmniMATH} for highly difficult, multi-branched advanced mathematical theorems. We also report zero-shot guided performance on subsets like \textbf{AIME}, \textbf{AMC}, \textbf{College Math}, and \textbf{Minerva Math}.

\textbf{SciBench \citep{wang2023scibench} \& GPQA \citep{rein2023gpqa}:} Used for out-of-domain evaluation on complex science reasoning. We uniformly sampled $1,000$ physics and chemistry data points from SciBench for domain adaptation. 

\section{Baselines and Comparison Methods}
\label{appendix_baselines}

We benchmark NS-PRM against a rigorous suite of state-of-the-art proprietary and open-weight process reward models and verification systems:

\begin{itemize}
    \item \textbf{Math-Shepherd-7B \citep{wang2023mathshepherd}:} A prominent process reward model initialized from Mistral/Llama weights, trained using automatically generated step-level annotations via Monte Carlo Tree Search.
    \item \textbf{Math-PSA-7B:} A process reward model leveraging step-aware learning paradigms to improve upon standard dense reward modeling.
    \item \textbf{RLHFlow Models \citep{dong2024rlhflow}:} Recent verification models trained on iterative reinforcement learning from human/AI feedback, built upon the Mistral-8B and DeepSeek-Math-8B architectures. 
    \item \textbf{Llemma-PRM800K-7B \citep{azerbayev2023llemma}:} The Llemma-7B foundational math model directly fine-tuned on the step-level verification annotations provided in the PRM800K open dataset.
    \item \textbf{Skywork-PRM-7B \citep{skywork2024}:} A state-of-the-art open-source PRM trained to evaluate step-by-step scientific and mathematical rationale.
    \item \textbf{ReasonEval-7B \citep{reasoneval2024}:} A dedicated reward model designed to output detailed evaluations of the intrinsic reasoning capability in autoregressive chain-of-thought traces.
    \item \textbf{Qwen2.5 Series \citep{yang2024qwen2}:} We utilize both the Qwen2.5-Math-7B and the highly optimized Qwen2.5-Math-PRM-7B as our direct architectural baselines.
    \item \textbf{R-PRM-7B \citep{she2025rprm}:} A highly recent robustness-focused process reward model trained with both Supervised Fine-Tuning (SFT) and Direct Preference Optimization (DPO) to resist adversarial mathematical perturbations.
\end{itemize}

\end{document}